\documentclass{bmvc2k}

\usepackage{amssymb}
\usepackage{gensymb}
\usepackage{multirow}
\usepackage{commath}

\title{A Simple Approach to Image Tilt Correction with Self-Attention MobileNet for Smartphones}

\addauthor{Siddhant Garg}{siddhantgarg@umass.edu}{1}
\addauthor{Debi Prasanna Mohanty}{debi.m@samsung.com}{2}
\addauthor{Siva Prasad Thota}{siva.prasad@samsung.com}{2}
\addauthor{Sukumar Moharana}{msukumar@samsung.com}{2}

\addinstitution{
 University of Massachusetts\\
 Amherst, USA
}
\addinstitution{
 On-Device AI\\
 Samsung Research,\\
 Bengaluru, India
}

\runninghead{Siddhant et al}{Self-Attention MobileNet for Image Tilt Correction}

\def\etal{\emph{et al}\bmvaOneDot}

\begin{document}

\maketitle

\begin{abstract}
Main contributions of our work are two-fold. First, we present a Self-Attention MobileNet, called SA-MobileNet Network that can model long-range dependencies between the image features instead of processing the local region as done by standard convolutional kernels. 
SA-MobileNet contains self-attention modules integrated with the inverted bottleneck blocks of the MobileNetV3 model which results in modeling of both channel-wise attention and spatial attention of the image features and at the same time introduce a novel self-attention architecture for low-resource devices. Secondly, we propose a novel training pipeline for the task of image tilt detection. We treat this problem in a multi-label scenario where we predict multiple angles for a tilted input image in a narrow interval of range $1\degree$ or $2\degree$, depending on the dataset used. 
With the combination of our novel approach and the architecture, we present state-of-the-art results on detecting the image tilt angle on mobile devices as compared to the MobileNetV3~\cite{mobilenetv3} model. Finally, we establish that SA-MobileNet is more accurate than MobileNetV3 on SUN397~\cite{sun397}, NYU-V1~\cite{nyuv1} and ADE20K~\cite{ade20k} datasets by 6.42\%, 10.51\%, and 9.09\% points respectively and faster by at least 4 milliseconds on Snapdragon 750 Octa core.
\end{abstract}
\begin{figure}
\centering
\begin{tabular}{ccc}
\bmvaHangBox{\fbox{\includegraphics[width=17mm]{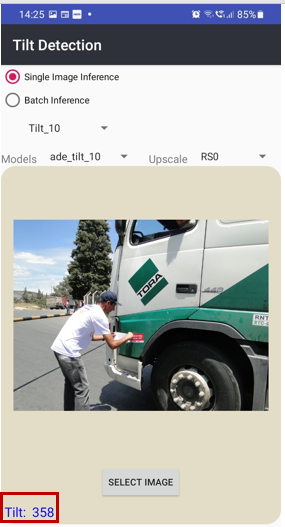}\hspace{2mm}\includegraphics[width=17mm]{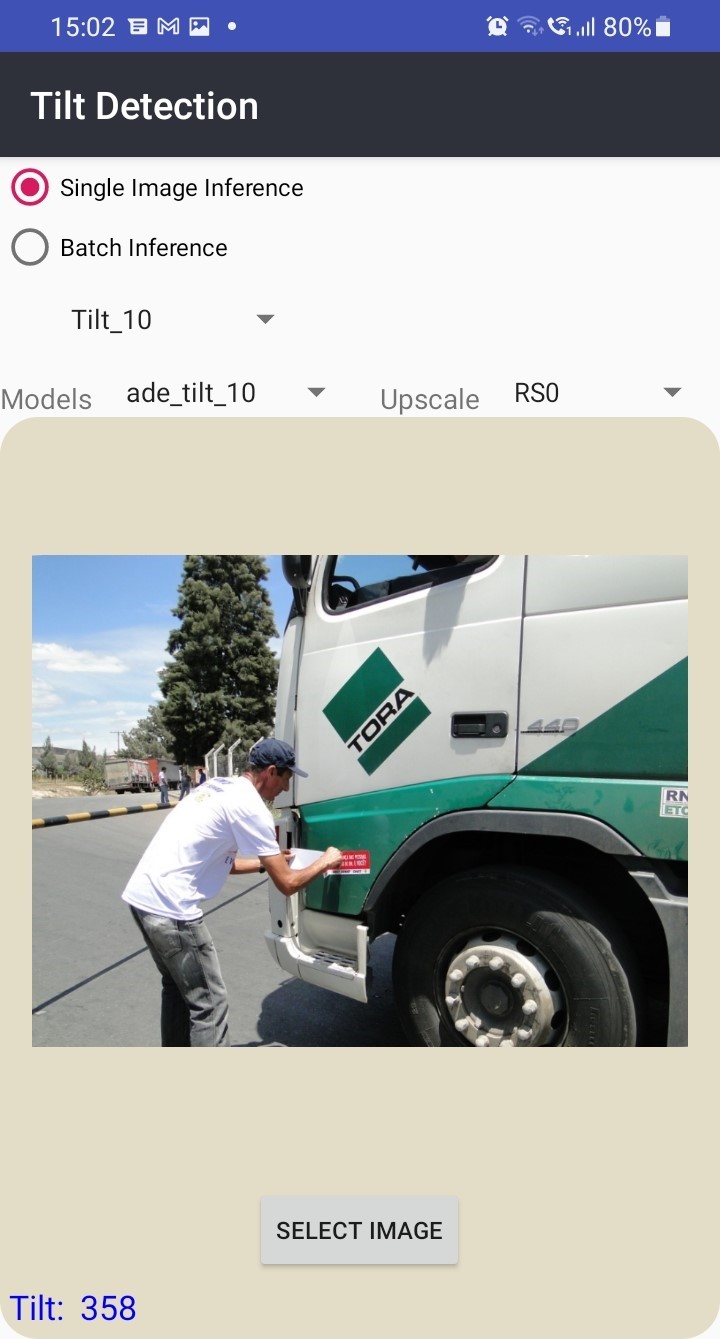}}}&
\bmvaHangBox{\fbox{\includegraphics[width=17mm]{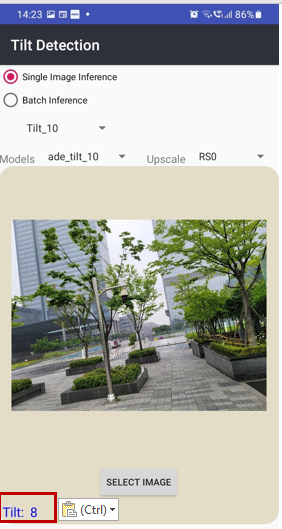}\hspace{2mm}\includegraphics[width=17mm]{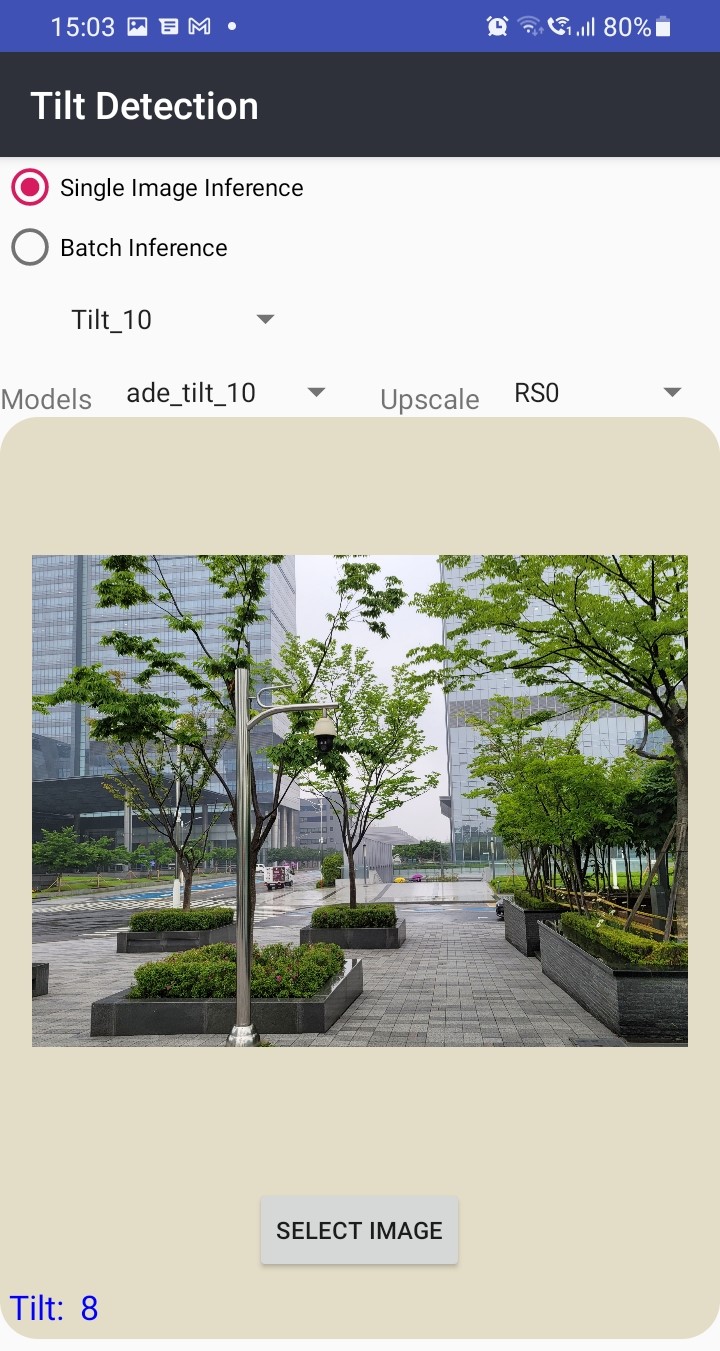}}}&
\bmvaHangBox{\fbox{\includegraphics[width=17mm]{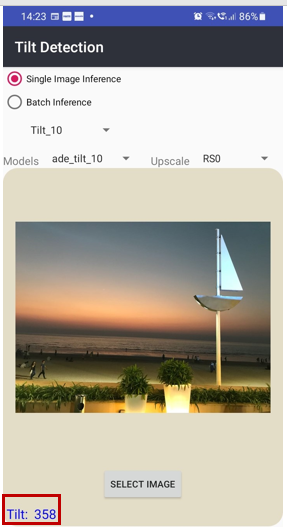}\hspace{2mm}\includegraphics[width=17mm]{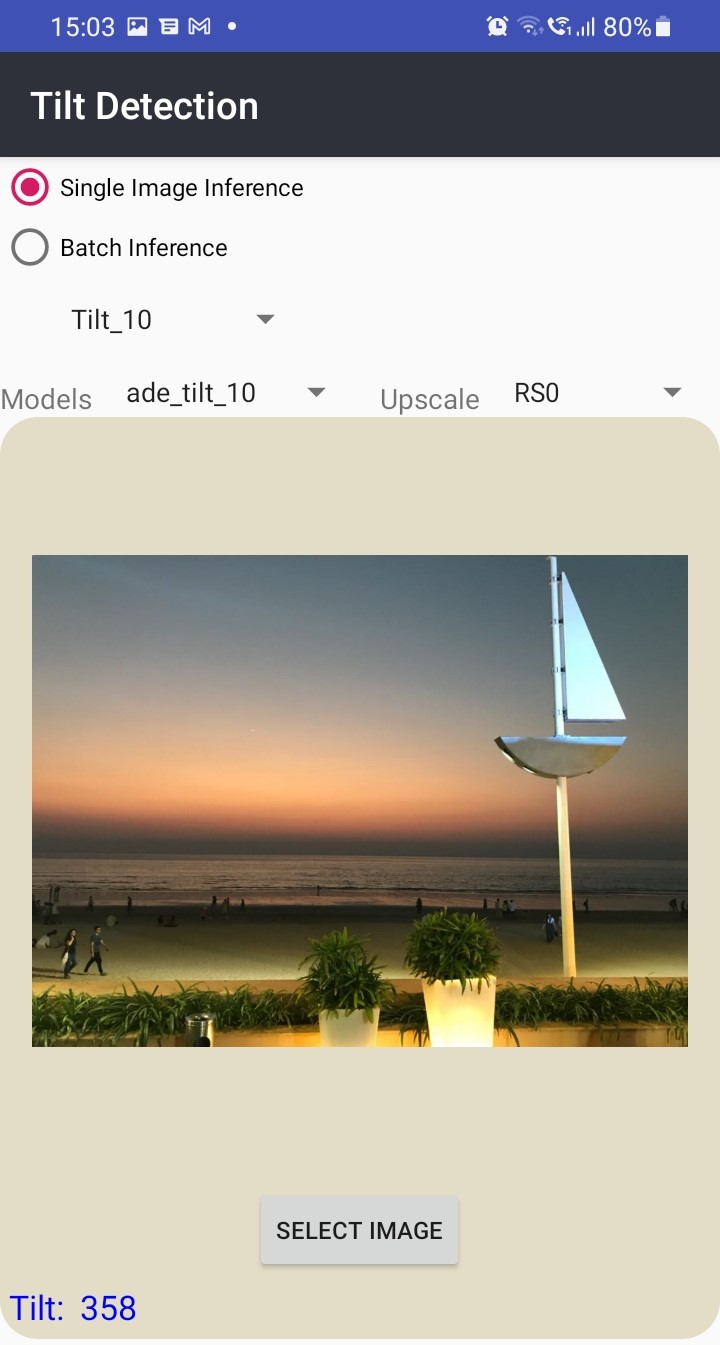}}}\\
\small{(a) Image is rotated $2\degree$ anti-} & \small{(b) Image is rotated $8\degree$ clock-} & \small{(c) Image is rotated $2\degree$ anti-} \\ 
\small{clockwise to align verti-} & \small{wise to make it upright.} & \small{clockwise to make the} \\
\small{cal lines on the truck} & \small{} & \small{horizon horizontal.}
\end{tabular}
\caption{Results on Image Tilt Correction in Real-Time. Proposed model is able to detect large as well as finer tilt angles. The predicted tilt angles are shown in a red box at bottom-left area of the images. Tilt angle value inside the box implies that image is tilted by that value in anti-clockwise direction from the upright orientation.}
\label{fig:results}
\end{figure}
\section{Introduction}
\label{sec:intro}
Smartphones have become the most convenient way to capture high-quality photos and videos. 
Mobile phone cameras have evolved over the past years with both hardware as well as software improvements with AI-enabled technologies that allow the users to take extremely high-resolution images. But many of us are not professional photographers and usually take images that are slightly skewed from the exact upright orientation. This reduces the aesthetic quality of images and people want that their holiday snapshots are of the highest quality. 
Professional photographers use softwares like Lightroom or Photoshop to straighten the tilted vertical or horizontal lines. 

We present an On-Device AI solution for automatic tilt angle detection of smartphone images and correct their orientations with a click to improve the overall picture quality. The proposed model is able to make inferences using mobile CPUs or GPUs with low latency values and at the same time respect the privacy of the user by removing the need to upload images to a server for processing. Currently, MobileNetV3~\cite{mobilenetv3} networks are the most popular lightweight models for mobile devices for many computer vision tasks like image classification or object detection. 

In this paper, we are proposing Spatial Self-Attention Modules that can learn long-range dependencies and global context within the input images. Furthermore, to enable on-device inference on resource-limited devices, we integrated these modules with the Inverted Bottleneck blocks \cite{sandler2018mobilenetv2} of MobileNetV3 to give us a novel neural network architecture for mobile devices called \textbf{Self-Attention MobileNet} or SA-MobileNet. 
The proposed network is able to learn the spatial information and overcomes the limitations of traditional convolutional kernels that only looks for different features in an image and not their relative positioning.

We are also proposing a simple yet effective training approach, described in Section \ref{sec:imagetilt}, to handle the image tilt detection problem that scales to variety of image datasets containing natural, or indoor/outdoor images.
The combination of the Self-Attention MobileNet and proposed training approach gives us state-of-the-art results for detecting image tilt for mobile devices in real-time.

Therefore, our two main contributions are:
\begin{itemize}
    \item Self-Attention MobileNet for mobile/IoT devices for real-time inference.  
\vspace{-8pt}
    \item A novel approach to tackle fine-grained Image Tilt Detection problem.
\end{itemize} 
\vspace{-11pt}
\section{Related Works}
\label{sec:related}
\textbf{Image Tilt Angle Detection} is a long-standing problem. Before the advent of deep learning, low-level image features were used to detect the upright image orientation like Ciocca \etal  ~\cite{ciocca2015lpb} who used LPB-based image features and logistic regression for this problem. But when the deep learning models for computer vision became popular, researchers started using high-level image features~\cite{vanishingnonmanhatten,wildhorizon,hold2018perceptual}. Fischer \etal~\cite{orientfisher} used AlexNet~\cite{krizhevsky2012alexnet} to regress the exact orientation angle of tilted images but there angle error was high($\approx20\degree$) for complete $360\degree$ range of image orientations. Applying CNNs for coarse-angle estimation and fuzzy logic for precise angle estimation on edge pixels~\cite{reshmalakshmi2017fuzzyedge,prince2019stepfuz} was recently employed to take into account the ambiguity and uncertainty in image orientations. Horizon lines and vanishing points are also used as cues for optimal image tilt detection but these methods are generally limited to outdoor images with a clear horizon line \cite{wildhorizon, fefilatyev2006horizon} whereas our work addresses natural images in diverse environments.

Digital camera parameters from accelerometer data are also used to rectify the image orientations. Do \etal\cite{spatialrectifier} proposed \textit{spatial rectifier} with ResNet-18 \cite{resnet} backbone network for surface normal estimation of indoor images. G Olmschenk \etal~\cite{pitchroll2d} proposed an InceptionNet \cite{inception} style architecture for estimating the pitch and roll of the camera from a single 2D image. Xian \etal\cite{uprightnet} used surface geometry to determine surface normals for estimating 2DoF~\cite{son2011twodof} camera orientations. But all of the above methods use neural networks that need high computational resources and it is not possible to deploy them in mobile/IoT devices. 

\textbf{Self-Attention} has also gained a lot of popularity in recent years. It has quickly become state-of-the-art baselines for many NLP tasks~\cite{waswani2017attention, parikh2016decomposable, cheng2016long} and after that it has also became popular in many computer vision tasks like Image Captioning~\cite{showattendtell}, Image Question Answering~\cite{imageattqa}, and Object Detection~\cite{detectiontransformer}. Self-Attention modules can model long-term dependencies and overcome the limitation of convolutional kernels which operate in a local neighbourhood~\cite{sagan}. BAM~\cite{park2018bam}, CBAM~\cite{cbam}, and ULSAM~\cite{ulsam} are light-weight attention modules for spatial attention but they use pooling operations that result in loss of information of the feature maps. 

\section{Method}
\label{sec:method}
We now present the architecture of our proposed Self-Attention MobileNet model and the novel training pipeline for Image Tilt Detection in detail in this section. 
\begin{figure}
\centering
\bmvaHangBox{\includegraphics[width=13cm]{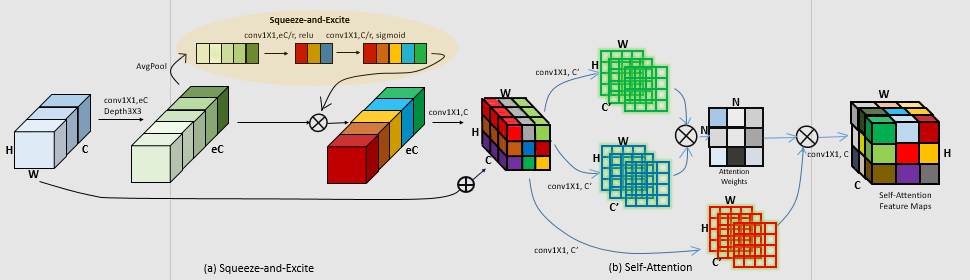}}
\caption{Complete Attention Module with Squeeze-and-Excite and Spatial Self-Attention. (a) $e$: expansion factor,  (b) $C`=C/r$, $r$: reduction ratio ($r=8)$. Best viewed in color.}
\label{fig:network}
\end{figure}

\subsection{Self-Attention Convolutional Module}
{\bf Squeeze-and-Excite}~\cite{hu2018squeeze}: 
The squeeze operation captures channel-wise dependencies in a feature map by using global average pooling operation along the channel dimension to aggregate information of each 2D feature map through a scalar value that results in a channel descriptor vector. During the excite operation, we learn channel-wise dependencies by passing the channel descriptor vector through a 2-layer neural network which outputs channel-wise attention weights. This attention vector is then multiplied by the input feature map to adaptively weight each channel and improve the representational power of the feature maps. 

{\bf Spatial Self-Attention}: Standard convolutional kernels process a local neighborhood of an image at a time because of the smaller receptive field sizes for the input feature maps. Although, the kernels with large receptive fields can cover more region but that will incur high computational costs because of the increased number of parameters and training time. Self-attention modules can complement convolutional layers by \textbf{learning global or long-range dependencies} between different image regions without much computational overhead. To learn spatial self-attention, we take the feature map vectors along the channel dimension and calculate their key, query, and value representations in a low-dimensional subspace. The dot product of all key vectors with every other query vector with the softmax function gives us the attention weights between all the regions of the image.
More specifically, after squeeze and excite operation, let the feature map be $\mathbf{F\in \mathbb{R}^{H\times W\times C}}$ and after expanding it along spatial dimensions we get the matrix $\mathbf{F\in \mathbb{R}^{N\times C}}$, which contains $N(=H\cdot W)$, $C$-dimensional vectors representing various image regions. We use learnable weight matrices $\mathbf{W_K\in \mathbb{R}^{C\times C`}}, \mathbf{W_Q\in \mathbb{R}^{C\times C`}}$, and $\mathbf{W_{v`}\in \mathbb{R}^{C\times C`}}$ to calculate key($\mathbf{K}$), query($\mathbf{Q}$) and value($\mathbf{v`}$) vectors respectively.
\begin{figure}
\centering
\begin{tabular}{cccc}
\bmvaHangBox{\includegraphics[width=2.6cm]{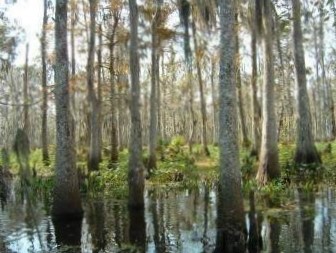}}&
\bmvaHangBox{\includegraphics[width=2.6cm]{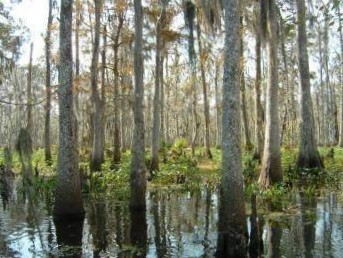}}&
\bmvaHangBox{\includegraphics[width=2.6cm]{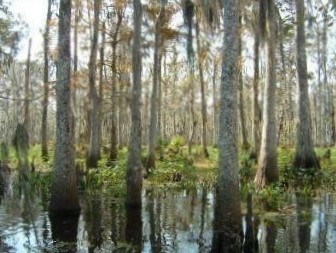}}&\bmvaHangBox{\includegraphics[width=2.6cm]{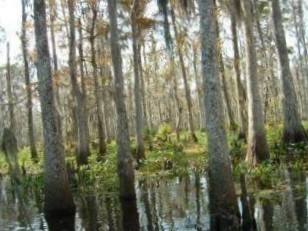}}\\
\small{(a) $-1\degree$} & \small{(b) $0\degree$} & \small{(c) $1\degree$} & \small{(d) $-5\degree$}\\
\bmvaHangBox{\includegraphics[width=2.6cm]{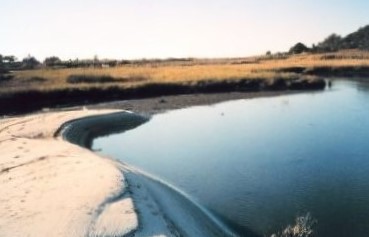}}&
\bmvaHangBox{\includegraphics[width=2.6cm]{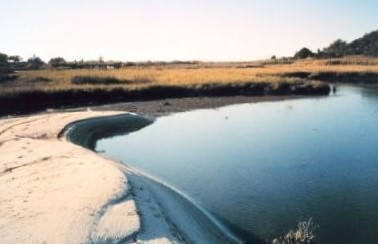}}&
\bmvaHangBox{\includegraphics[width=2.6cm]{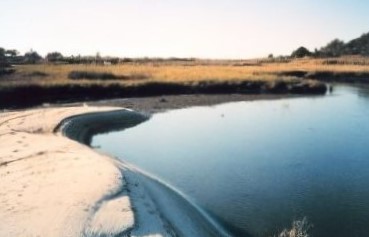}}&\bmvaHangBox{\includegraphics[width=2.6cm]{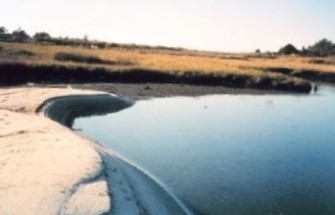}}\\
\small{(e) $-1\degree$} & \small{(f) $0\degree$} & \small{(g) $1\degree$} & \small{(h) $5\degree$}
\end{tabular}
\caption{Images (a), (b), (c) and (e), (f), (g) can be perceived as upright because of very small deviations from the upright orientation. But the Fig.(d) needs to be rotated $5\degree$ clockwise and Fig.(h) by $5\degree$ anti-clockwise to make them upright.}
\label{fig:intro}
\end{figure}
Here $C` = C / r$, $r > 1$, $r$ is the reduction ratio which is used to decrease the dimensions of the vectors and calculate attention weights and values in a low-dimensional subspace.
Let $\mathbf{F^{N\times C}}\equiv \mathbf{F}$ 
\begin{align}
    \mathbf{K} = \mathbf{F \cdot W_K}, \quad \mathbf{Q} &= \mathbf{F \cdot W_Q}, \quad \mathbf{v`} = \mathbf{F \cdot W_{v`}} \\
    \mathbf{\beta^{N\times N}} = \mathbf{Q \cdot K^T} \implies a_{ij} &= \frac{\exp(\beta_{ij})}{\sum_{j=1}^{N}\exp(\beta_{ij})}, i,j=1,\dots,N \\
    \mathbf{V^{N\times C`}} &= \mathbf{a} \cdot \mathbf{v`} \label{V}
\end{align}
Here, $\mathbf{a\in \mathbb{R}^{N\times N}}$ is the self-attention matrix, and $a_{ij}$ is the attention weight that region $i$ puts on region $j$. $\mathbf{V\in \mathbb{R}^{N\times C`}}$in Eq.\ref{V} is a value matrix in the low-dimensional subspace and we use $\mathbf{W_V\in \mathbb{R}^{C`\times C}}$ to project it into the original subspace  to get the self-attention feature maps, $\mathbf{S\in \mathbb{R}^{N\times C}}$. Finally, we will add a residual connection to get the final output $\mathbf{\Tilde{F}^{N\times C}}$.
\begin{align}
    \mathbf{S} &= \mathbf{V} \cdot \mathbf{W_{V}} \label{S}\\
    \mathbf{\Tilde{F}} &= \mathbf{F} + \alpha \mathbf{S} \label{ft}
\end{align}
where $\alpha$ is a trainable scalar parameter initialized to $0$. Eq.\ref{ft} implies that the model first learns image features around the local neighbourhood and then gradually moves on to learn global dependencies~\cite{sagan}. The combination of squeeze and excite operation and spatial self-attention gives us the feature maps that are rich in content as well as contextual information. 

\begin{figure}
\centering
\begin{tabular}{cc}
\bmvaHangBox{\fbox{\includegraphics[width=14mm]{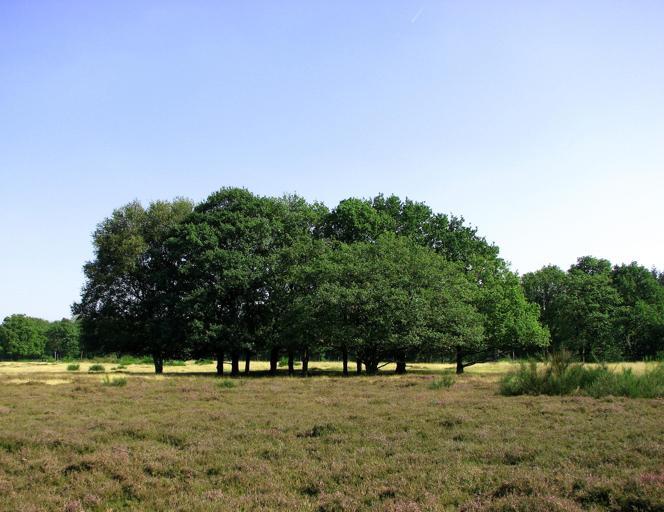}\hspace{1mm}\includegraphics[width=14mm]{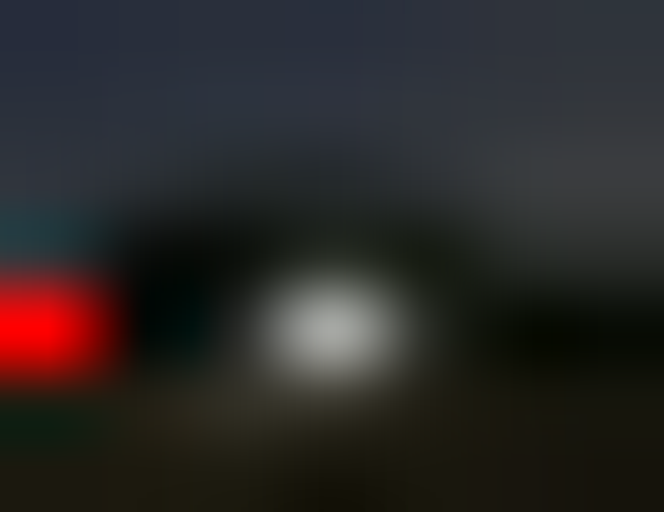}\hspace{1mm}\includegraphics[width=14mm]{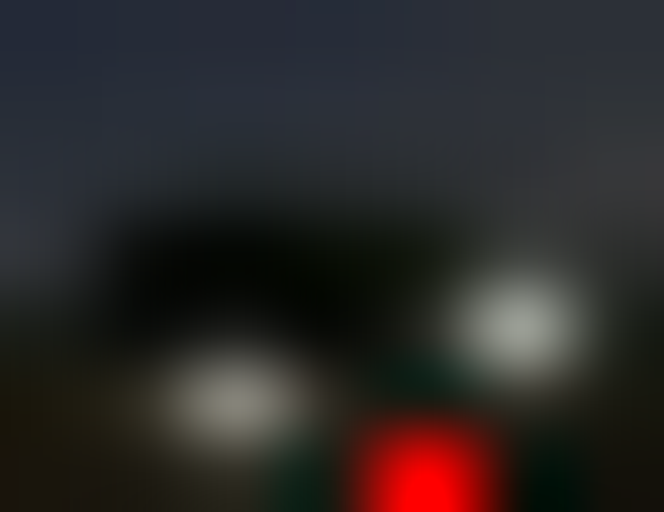}\hspace{1mm}\includegraphics[width=14mm]{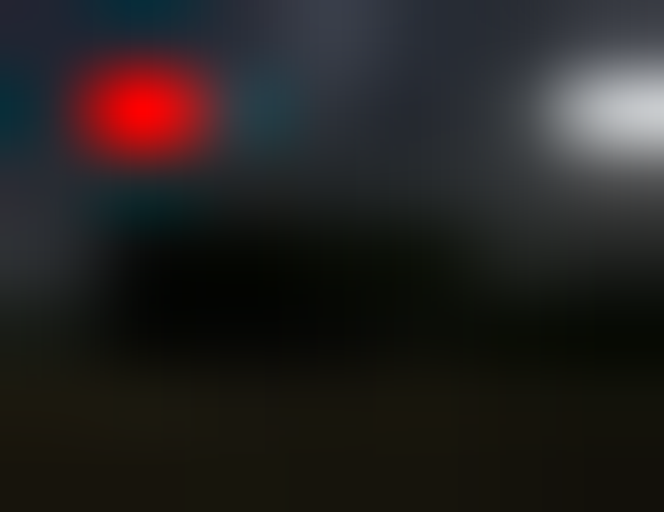}}}&
\bmvaHangBox{\fbox{\includegraphics[width=14mm]{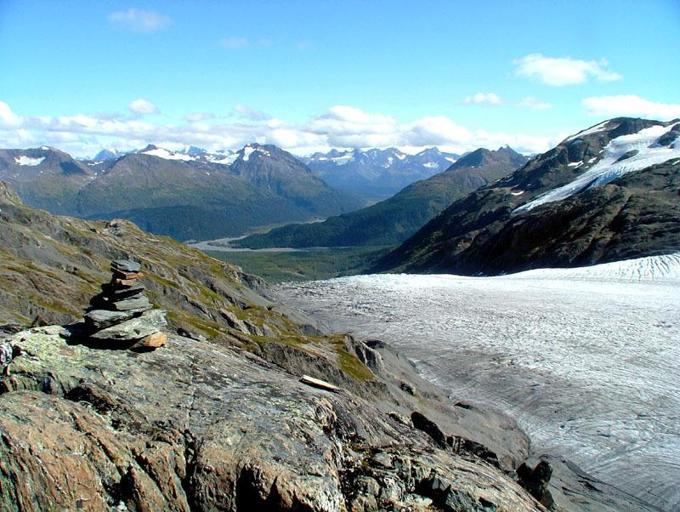}\hspace{1mm}\includegraphics[width=14mm]{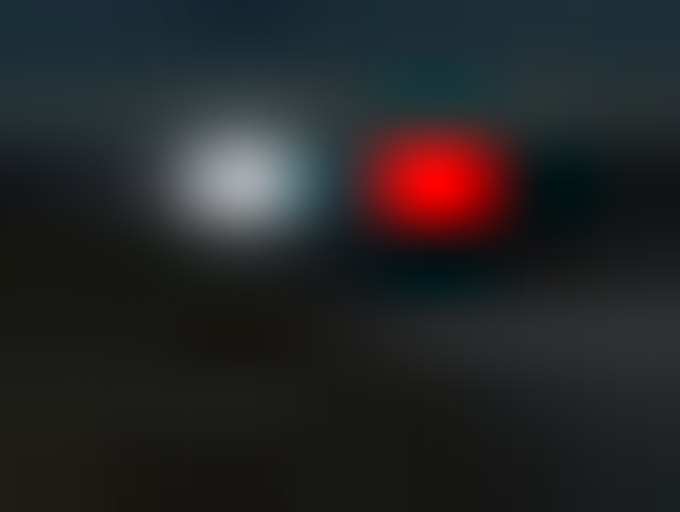}\hspace{1mm}\includegraphics[width=14mm]{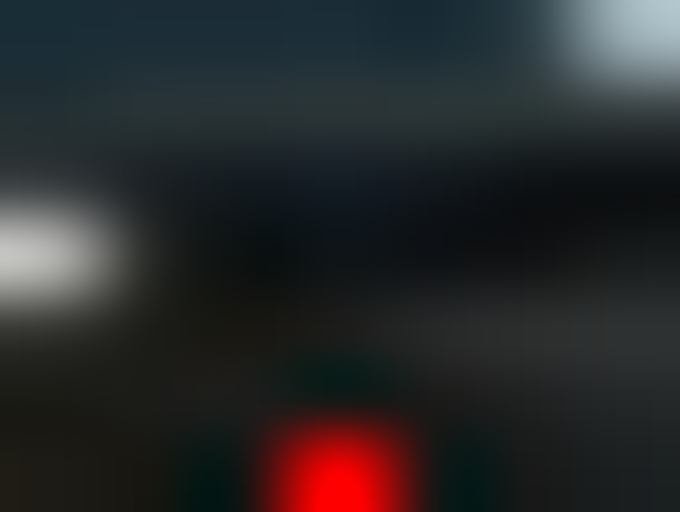}\hspace{1mm}\includegraphics[width=14mm]{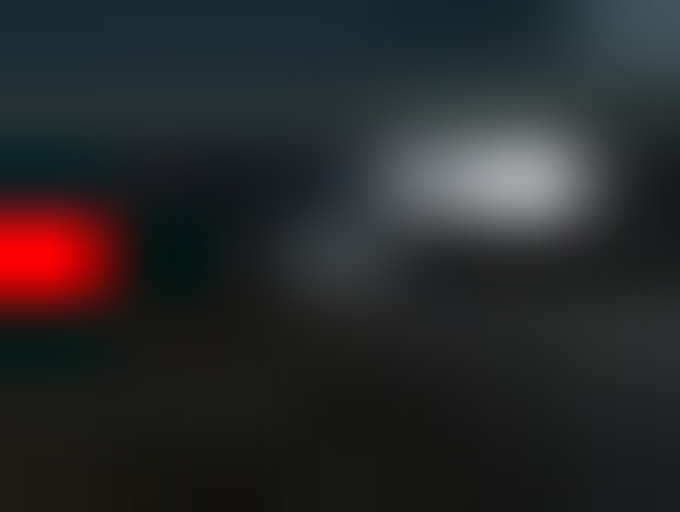}}}\\
{\scriptsize $1 \hspace{14mm} 1.a \hspace{12mm} 1.b\hspace{14mm} 1.c$} & {\scriptsize $4 \hspace{14mm} 4.a \hspace{12mm} 4.b\hspace{14mm} 4.c$} \\
\bmvaHangBox{\fbox{\includegraphics[width=14mm]{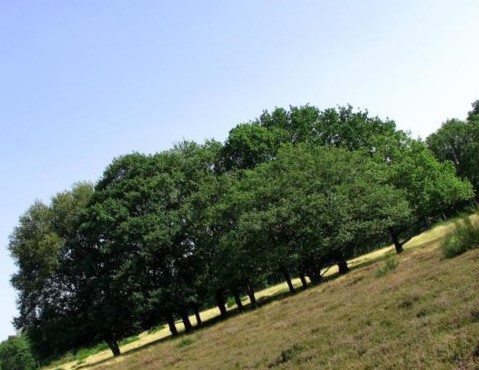}\hspace{1mm}\includegraphics[width=14mm]{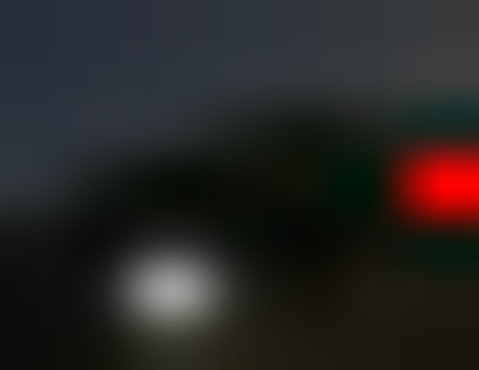}\hspace{1mm}\includegraphics[width=14mm]{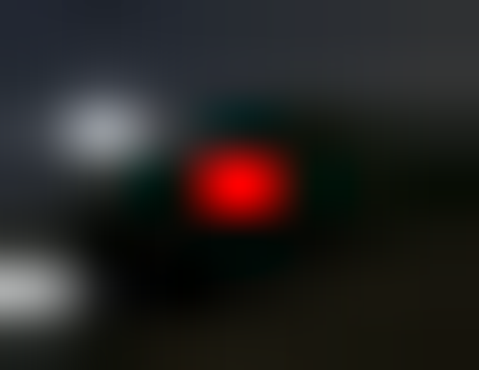}\hspace{1mm}\includegraphics[width=14mm]{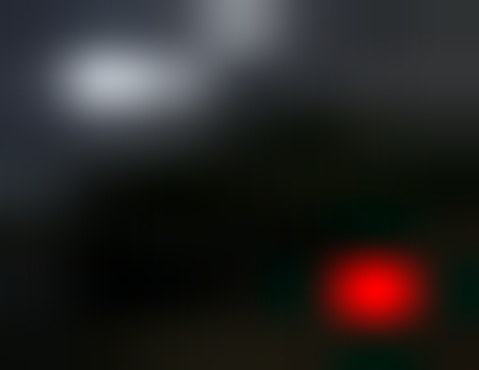}}}&
\bmvaHangBox{\fbox{\includegraphics[width=14mm]{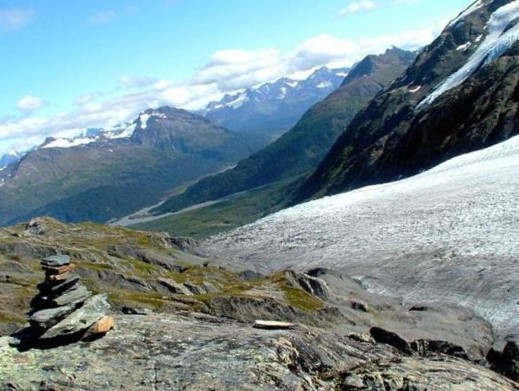}\hspace{1mm}\includegraphics[width=14mm]{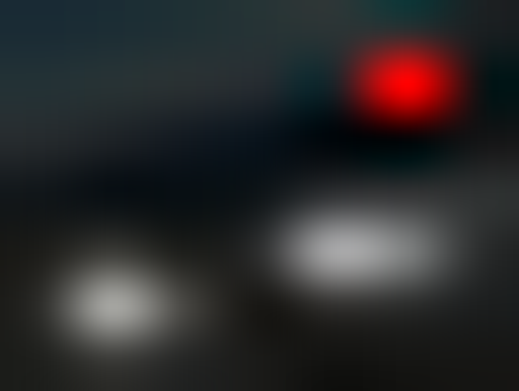}\hspace{1mm}\includegraphics[width=14mm]{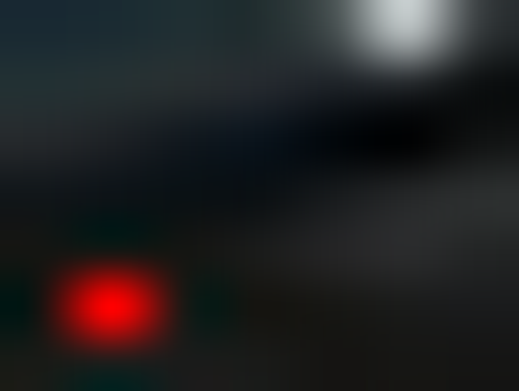}\hspace{1mm}\includegraphics[width=14mm]{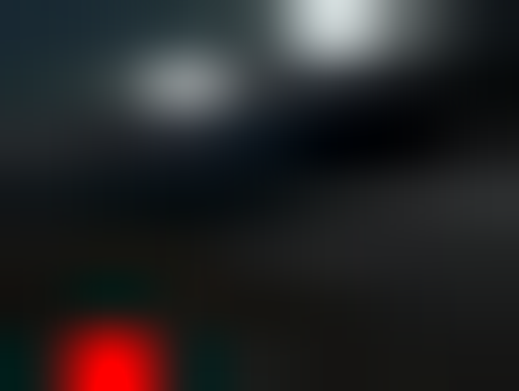}}}\\
{\scriptsize $2 \hspace{14mm} 2.a \hspace{12mm} 2.b\hspace{14mm} 2.c$} & {\scriptsize $5 \hspace{14mm} 5.a \hspace{12mm} 5.b\hspace{14mm} 5.c$} \\
\bmvaHangBox{\fbox{\includegraphics[width=14mm]{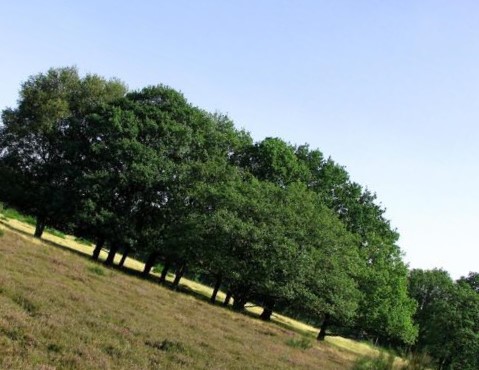}\hspace{1mm}\includegraphics[width=14mm]{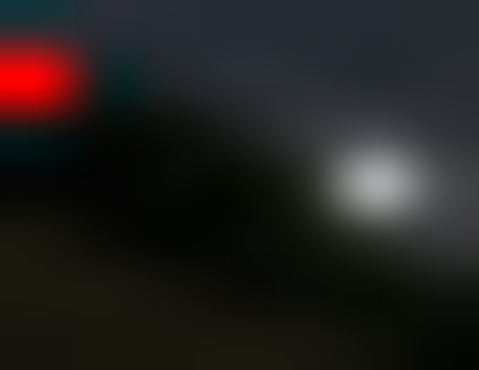}\hspace{1mm}\includegraphics[width=14mm]{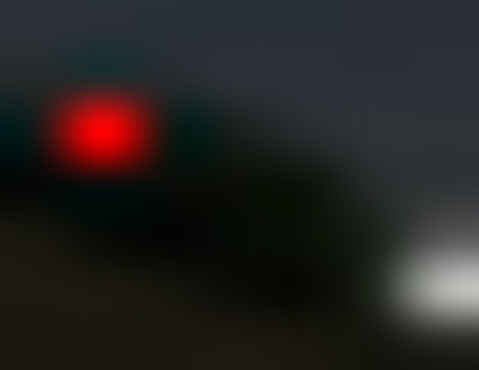}\hspace{1mm}\includegraphics[width=14mm]{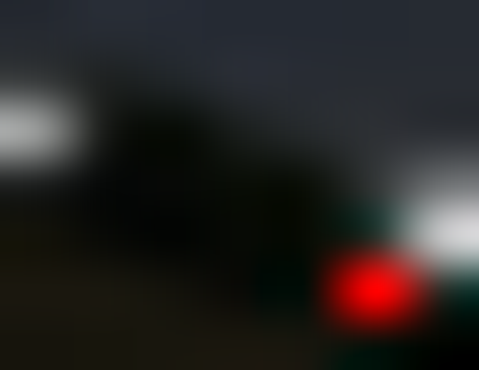}}}&
\bmvaHangBox{\fbox{\includegraphics[width=14mm]{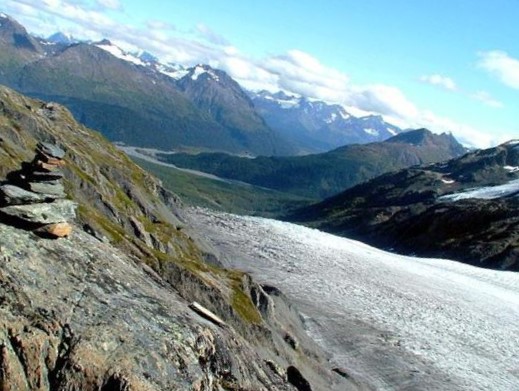}\hspace{1mm}\includegraphics[width=14mm]{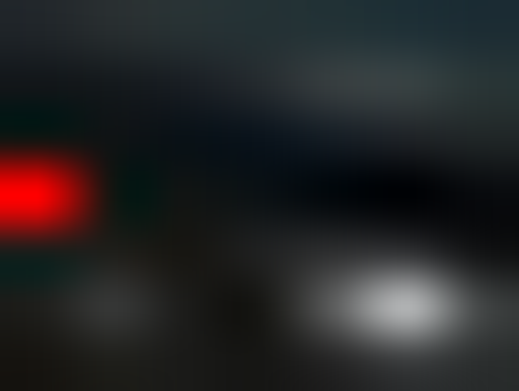}\hspace{1mm}\includegraphics[width=14mm]{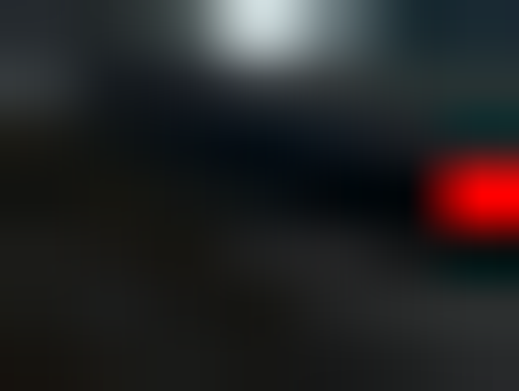}\hspace{1mm}\includegraphics[width=14mm]{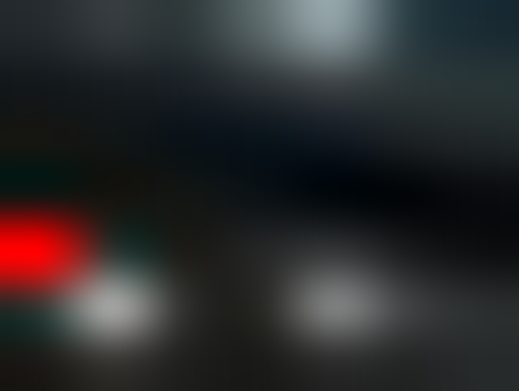}}}\\
{\scriptsize $3 \hspace{14mm} 3.a \hspace{12mm} 3.b\hspace{14mm} 3.c$} & {\scriptsize $6 \hspace{14mm} 6.a \hspace{12mm} 6.b\hspace{14mm} 6.c$} \\
\small{SET-A}&\small{SET-B}  
\end{tabular}
\caption{In the heatmaps, the red region indicates the query region and the white regions represent the points where the query region focuses its attention. Figures. 1.a, 2.a, and 3.a indicates that the query region on the horizon attends to other regions on the horizon. For the SET-B, the input images do not have straight lines but the model is able to understand the general upright orientation. In figures 4.c and 6.a, the query region on ground is attending to other ground regions. Similarly, in figures 4.a and 6.b the query region in sky is attending to another point in the sky. In figures 1.c, 2.c, and 3.c and figures 4.c, 5.c, and 6.c, the query region is attending to a far location indicating the learning of long-range dependencies.}
\label{fig:heatmaps}
\end{figure}

The learning of global long-range dependencies and relative positioning of different image regions improved the model predictions on image tilt detection tasks. From the heatmaps, in Fig.\ref{fig:heatmaps}, we can see that the query (red) region is able to shift its attention with the image orientation. Specifically, in Fig.\ref{fig:heatmaps}\textcolor{red}{-[1.a, 2.a, 3.a]} (left column), the query (red) region on the horizon was able to focus on the other horizon points despite different orientations of the same image. In Fig.\ref{fig:heatmaps}\textcolor{red}{-SET-B} (right column), there is no clear horizon line but the query regions are able to attend to relevant regions for tilt detection. 
For example, query region on the ground in heatmaps-\textcolor{red}{[4.b, 5.b]} is attending to other points on the ground and query region in Fig.\ref{fig:heatmaps}\textcolor{red}{-[4.a, 6.b]} in sky is focusing on other points above the skyline. 
Also note that query regions in Fig.\ref{fig:heatmaps}\textcolor{red}{-[1.c, 2.c, 3.c]} (left column) are able to attend to far locations which indicates learning of long-range dependencies and \textbf{spatial information}. Therefore, for detecting image tilt angle, the neural network model needs to \textbf{learn the relative positioning of the image pixels} to differentiate between various distinct image orientations. 
\subsection{Image Tilt Detection}
\label{sec:imagetilt}
\subsubsection{Motivation and Problem Modeling}
\label{motivation}
Although, the intuitive approach for tackling this problem seems to be regression where we use a Deep CNN to extract image features and minimize the angular distance between the ground truth angle and the predicted angle. But training deep neural networks for regression tasks is difficult~\cite{reg1} and when we enter the domain of light-weight models, like MobileNetV3 or SA-MobileNet, with few parameters as compared to ResNet-50~\cite{resnet} or VGG-16~\cite{vgg} networks, it becomes extremely difficult to achieve good results on regression tasks. In contrast to regression, deep neural networks achieve highly accurate results on classification tasks but we cannot model the image tilt detection as a single label classification problem due to a variety of reasons. If the prediction is $1\degree-2\degree$ off from the true label, the network will penalize it equally as it would if the prediction is off by $10\degree-20\degree$ or more. This might not be necessary because $1\degree$ or $2\degree$ variation in the image tilt angle might not be significant. Therefore, \textbf{we train the model to predict multiple angles within a narrow interval} of the ground truth tilt angle and penalize only those values that are outside this narrow range.
\subsubsection{Training Pipeline}
We model the problem of Image Tilt Detection in a multi-label scenario where we train our proposed neural network model with a tilted input image and the corresponding ground truth label vector with multiple labels of tilt angles in a narrow interval of either $\pm 1\degree$ or $\pm 2\degree$ depending on the dataset quality. 
Images in the training dataset were assumed to be upright and assigned $0\degree$ ground truth label by default. Every image is rotated by each angle from $0\degree, 1\degree, \dots, 359\degree$ and center-cropped before giving to the model as input. 

To enable multi-label training for predictions of image tilt angles within the narrow interval, the ground truth label vectors are constructed as 360-dimensional vectors with value of $1$ at indices $G-I, \dots, G-1, G, G+1, \dots, G+I$, and value of $0$ at rest of the indices. $G$ and $I$ are the ground truth image tilt angle and length of the narrow interval respectively. Note that the ground truth label vector is cyclic. If $G+x\geq 360$, for some $0\leq x\leq I$, then we set the value 1 at index $(G+x)\%360$. Similarly, if $G-x<0$, then we set 1 at index $G-x+360$. 
Last layer of the model is a 360-dimensional layer, representing all the integer angles, with sigmoid activation function that holds the prediction scores of the tilt angles for a given input image. We use Binary Cross Entropy loss function to calculate the loss between given ground truth label vector and the predicted outputs. 
\begin{align}
    \mathbf{\mathcal{L}(y, p)} = -\frac{1}{D}\Bigg[\sum_{i=1}^{D}y_{i}\log p_i + (1-y_i)\log(1-p_i)\Bigg]
\end{align}
Here $D=360$ and $\mathbf{y}$ and $\mathbf{p}$ are groundtruth and the predicted vectors respectively.

\section{Experiments}
\label{sec:experiments}
\subsubsection{Model Prediction}
For the final prediction, we take the highest scoring label, in the last layer, and if more than two labels have the highest scores, we simply take their average value. From our experiments, we saw that most of the times, when more than two labels have the highest score, it was $1.0$ and the corresponding labels were consecutive and belonged to set of narrow interval around the ground truth tilt angle. 
This implies that there is an \textbf{implicit correlation between the ground truth labeled angles} within the narrow interval that helps the model to determine the image orientation over the complete $360\degree$ range. Another advantage of using this method, over single-label classification, is that the network only penalizes those output values that are outside the  narrow interval around the ground truth tilt angle. Intuition is that if the ground truth image tilt angle is $45\degree$, then we do not want to penalize $44\degree$ or $46\degree$ prediction as much as we want to penalize $42\degree$ or $46\degree$ predictions because though they are close to the ground truth value they will not make image look upright.
\subsection{Model Architecture}
\label{sec:model}
For the MobileNetV3 model, the input image size is $224\times 224$ and the subsequent layers decrease the feature map size to $7\times 7$ in strides of $2$. We integrated Spatial Self-Attention Modules within MobileNetV3 to get Self-Attention MobileNet.
Spatial Self-Attention Modules were applied to the blocks of sizes $56\times 56$, $28\times 28$, $14\times 14$, and $7\times 7$. 
We selected these blocks because they are high-level image features which encodes meaningful image representations. Furthermore, the computational costs of calculating self-attention on these blocks was not very high because of the small feature map sizes.  
We also replaced the 1280-dimensional fully connected layer of MobileNetV3 with a 720-dimensional layer. This helped in reducing the MAdds that were added due to the introduction of spatial self-attention operations.
As a result, the proposed SA-MobileNet model, when converted to Tflite, came out {\bf faster by an average of 4ms} than the corresponding MobileNetV3 Tflite model when tested on Snapdragon 750 Octa core. All the convolutional kernels and fully connected layers were initialized from pretrained ImageNet~\cite{krizhevsky2012alexnet} weights apart from the newly added self-attention blocks that were initialized randomly.

The network was trained end-to-end using RMSprop~\cite{rms} optimizer with momentum $0.9$. The initial learning rate was $0.001$ and it was decayed using exponential learning rate schedule with $40k$ decay steps and $0.95$ decay rate.
\begin{table}
\begin{center}
\begin{tabular}{|l|c|c|}
    \hline
    \textbf{Previous Works} & \textbf{Accuracy} $(\%) \uparrow$ & \textbf{Angle Error} $(\degree) \downarrow$ \\
    \hline
    Ciocca \etal ~\cite{ciocca2015lpb} (LPB-based featutes) & 71.87 & - \\ \hline 
    CNN + Fuzzy Edge Detection & 85.21 & - \\ \hline
    Fischer \etal ~\cite{orientfisher} (AlexNet) & - & 21.23 \\ \hline
    Maji \etal ~\cite{maji2020deep} (Xception) & - & 7.89 \\ \hline 
    MobileNetV3 (baseline) & 85.97 & 5.06 \\ \hline
    ResNet-50 (baseline) & \textbf{93.67} & \textbf{3.98} \\ \hline
    SA-MobileNet (proposed) & 92.39 & 4.27 \\ \hline
\end{tabular}
\caption{Accuracies and angles errors of various baseline methods on SUN397 dataset. $\uparrow/\downarrow$ indicates that higher/lower is better respectively.}
\label{tab:previousworks}
\end{center}
\end{table}

\subsection{Datasets}
\label{sec:datasets}
\begin{figure}
\centering
\begin{tabular}{ccc}
\bmvaHangBox{\fbox{\includegraphics[width=18mm]{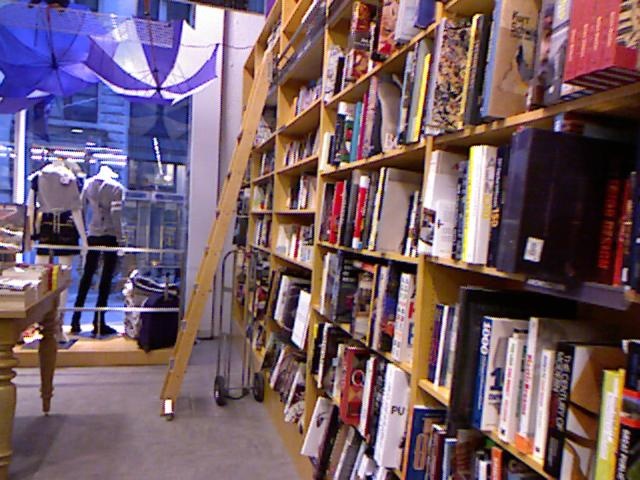}\hspace{1mm}\includegraphics[width=18mm]{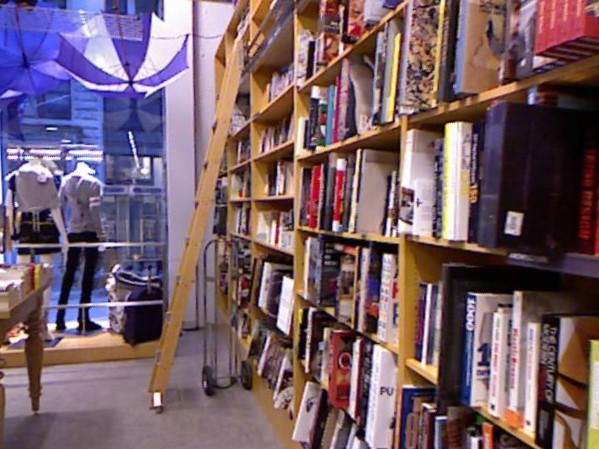}}}&
\bmvaHangBox{\fbox{\includegraphics[width=18mm]{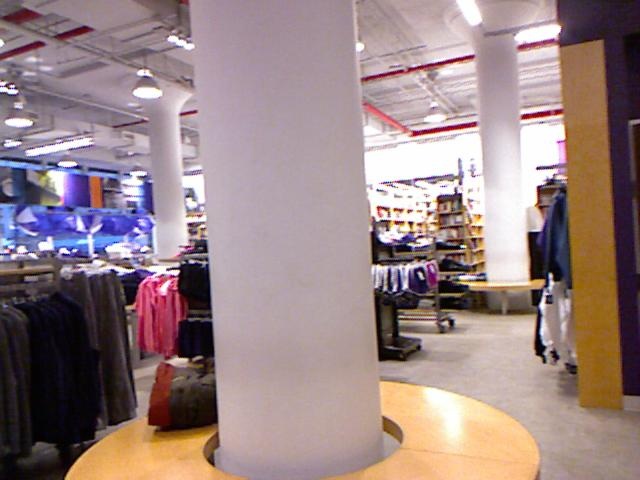}\hspace{1mm}\includegraphics[width=18mm]{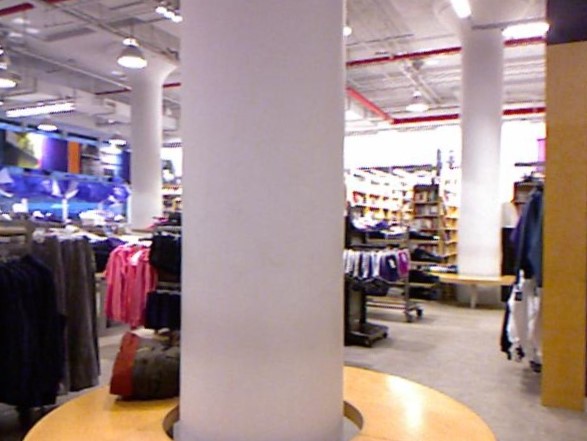}}}&
\bmvaHangBox{\fbox{\includegraphics[width=18mm]{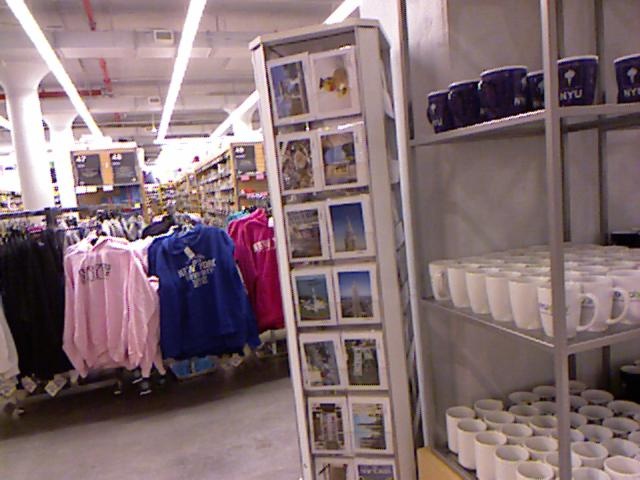}\hspace{1mm}\includegraphics[width=18mm]{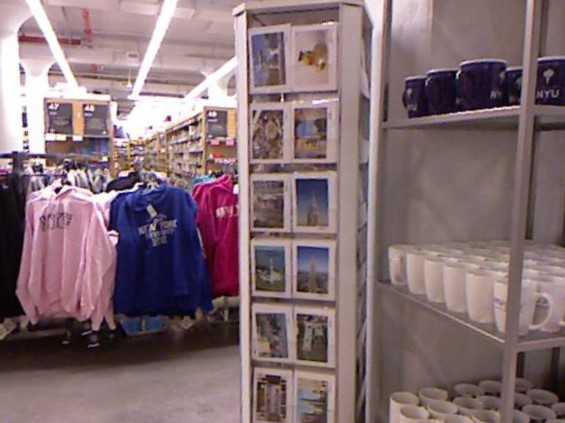}}}\\
\small{(a) $3\degree$ anticlockwise} & \small{(b) $4\degree$ clockwise} & \small{(c) $5\degree$ clockwise} 
\end{tabular}
\caption{Samples of images NYU-V1 dataset. Most the images(left) were skewed and had to be rotated to make them upright(right).}
\label{fig:nyu}
\end{figure}

We used publicly available SUN397~\cite{sun397}, ADE20K~\cite{ade20k}, and NYU-V1~\cite{nyuv1} datasets. SUN397 is a scene understanding dataset that contains 397 well-sampled categories of diverse scenes with $108,754$ distinct images. There are 10 train-test partitions for the dataset, and for our evaluation dataset, we took the union of images from all the test partitions that resulted in $15,691$ images. The remaining $92,793$ images were used for training. ADE20K dataset is another scene parsing dataset with $20,210$ images in the training set and $5,000$ images in the evaluation set. SUN397 and ADE20K datasets contain wide variety of natural images that may or may not contain straight vertical or horizontal lines. That is why we set the interval length, $I=2\degree$ while training on these two datasets. 
We also used NYU-V1 Depth dataset that contains frames from video sequences of various indoor scenes recorded from both RGB and Depth camera of Microsoft Kinect. 
Before using this dataset to train our model, we had to make the images upright because all the images were skewed as seen from Fig.\ref{fig:nyu}. We straightened the set of $2,282$ images and used $2000$ images for our training and $282$ images for testing. We split the data in such a way that frames from the same indoor scene does not come in both the splits. We set the interval length $I=1\degree$ because this dataset was manually annotated and we saw highly accurate results on the evaluation data. 

\subsection{Results}
\label{sec:results}

\begin{figure}
\centering
\begin{tabular}{cccc}
\bmvaHangBox{\includegraphics[width=2.88cm]{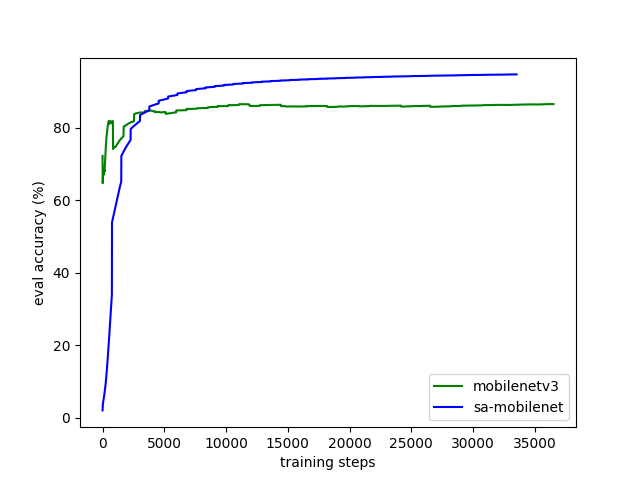}}&
\bmvaHangBox{\includegraphics[width=2.88cm]{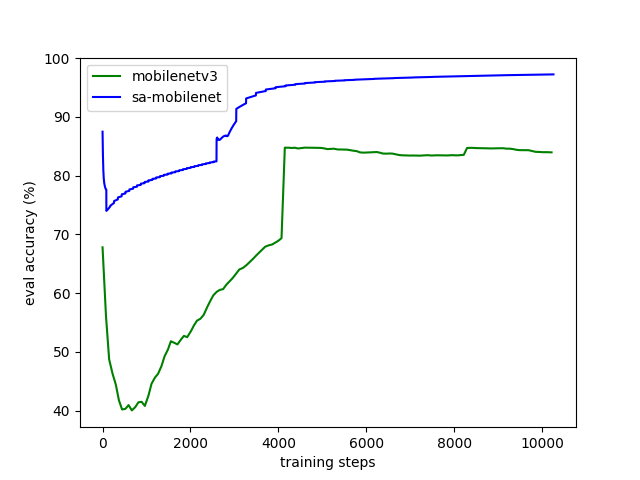}}&
\bmvaHangBox{\includegraphics[width=2.88cm]{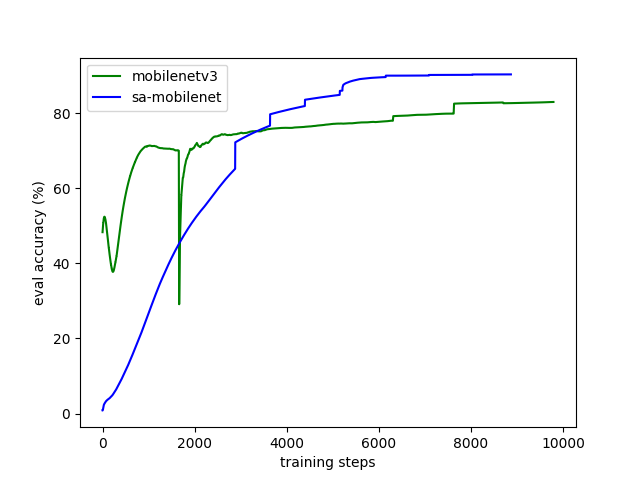}}& \bmvaHangBox{\includegraphics[width=2.88cm]{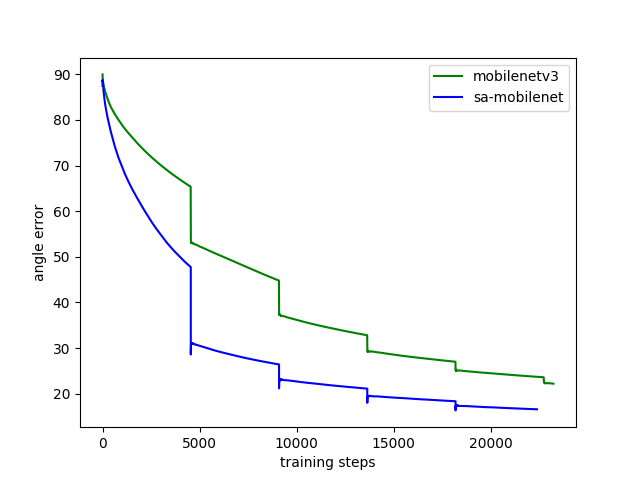}}\\
\small{(a) ADE20K} & \small{(b) NYU-V1} & \small{(c) SUN397} & \small{(d) Regression loss}
\end{tabular}
\caption{Figures (a), (b), and (c) contains evaluation accuracy plots over training steps for various datasets. The accuracy curve of SA-MobileNet (blue) model is above the accuracy curve of MobileNetV3 (green) model. Fig (d) contains the plots for regression losses calculated using angle error loss function (Eq.\ref{angleloss}) and trained on ADE20K dataset.}
\label{fig:plots}
\end{figure}

We train MobileNetV3 model as a baseline for mobile devices on SUN397, ADE20K and NYU-V1 dataset using the training approach, described in section \ref{sec:imagetilt}. The proposed SA-MobileNet consistently performs better in terms of detection accuracies and angle errors on all the evaluation datasets as seen from Fig.\ref{fig:plots} and Table.\ref{tab:accuracies}. We also trained MobileNetV3 and SA-MobileNet for regressing the image tilt angle, by using angle loss function, given by Eq.\ref{aloss} and \ref{angleloss}, and AdaDelta optimizer on ADE20k dataset. The SA-MobileNet model gives us lower angle error when compared to the MobileNetV3 model as shown in Fig \ref{fig:plots}\textcolor{red}{.d} and Table \textcolor{red}{3}.
\vspace{-10pt}
\begin{align}
e &= \abs{a_{true} - a_{pred}} \label{aloss}\\
\mathbf{\mathcal{L}_{angle}} &= min\{e, 360-e\} \label{angleloss}
\end{align}
Here $a_{true}$ and $a_{pred}$ are groundtruth and predicted angles respectively, with values ranging from $0\degree-to-359\degree$. We also use this loss function to calculate the angle errors of the model trained with the multi-label approach. 
\begin{table}[h!]
\begin{center}
\begin{tabular}{|c|c|c|c|c|c|c|}
    \hline
    \multirow{2}{*}{\textbf{Model}} & \multicolumn{2}{c|}{\textbf{NYU-V1}} & \multicolumn{2}{c|}{\textbf{ADE20K}} & \multicolumn{2}{c|}{\textbf{SUN397}}\\ \cline{2-7}
     & Acc$(\%)\uparrow$ & AE$\degree\downarrow$ & Acc$(\%)\uparrow$ & AE$\degree\downarrow$ & Acc$(\%)\uparrow$ & AE$\degree\downarrow$\\ \hline
     MobileNetV3 & 88.02 & 15.79 & 87.68 & 16.84 & 85.97 & 5.06\\ \hline
     ResNet-50 & 94.59 & 4.67 & \textbf{97.84} & \textbf{3.09} & \textbf{93.67} & \textbf{3.98}\\ \hline
     SA-MobileNet & \textbf{98.53} & \textbf{3.45} & 96.77 & 3.45 & 92.39 & 4.27\\ \hline
\end{tabular}
\caption{Evaluation accuracies and angle errors of the MobileNetV3, ResNet-50, and SA-MobileNet models on various datasets with the proposed tilt angle detection approach. Acc: Accuracy $(\%)$ and AE: Angle Errors($\degree$). $\uparrow/\downarrow$ indicates that higher/lower is better respectively.}
\label{tab:accuracies}
\end{center}
\begin{center}
\parbox{.5\linewidth}{
\centering
    \begin{tabular}{|c|c|c|}
    \hline
    \textbf{Model} & \textbf{Latency}$(\downarrow)$ & \textbf{Parameters$(\downarrow)$}\\
    & (milliseconds) & (millions)\\ 
    \hline
    MobileNetV3 & $79$ & \textbf{4.2} \\ \hline 
    SA-MobileNet & \textbf{75} & 4.5\\ \hline
    \end{tabular}
\label{tab:parameters}
\caption{Tflite models were tested on Snapdragon 750, Octa
core (2x 2.2 GHz, 6x 1.8 GHz) for latency measurements.}
}
\hfill
\parbox{.35\linewidth}{
\centering
\begin{tabular}{|c|c|}
    \hline
    \textbf{Model}&\textbf{Angle}\\ 
    & \textbf{Error$\degree$ $(\downarrow)$} \\ \hline
    MobileNetV3 & $21.07$ \\ \hline 
    SA-MobileNet & \textbf{15.53}\\ \hline
\end{tabular}
\label{tab:regtable}
\caption{Regresion loss on ADE20k dataset trained with angle loss function Eq.\ref{angleloss}.}
}
\end{center}
\end{table}
From Table.\ref{tab:accuracies} we see that the proposed SA-MobileNet model resulted in very low-angle errors on NYU-V1, ADE20K, and SUN397 dataset when compared with MobileNetV3 model. From the evaluation accuracy plots in Fig.\ref{fig:plots}\textcolor{red}{.a}, Fig.\ref{fig:plots}\textcolor{red}{.b}, Fig.\ref{fig:plots}\textcolor{red}{.c}, we can see that the accuracy curve of our model (blue) is above the curve of MobileNetV3 model (green) during training for all the datasets. In Fig.\ref{fig:plots}\textcolor{red}{.d}, we plot angle errors for both the models which were trained for regressing the exact orientation angle on ADE20K dataset. The SA-MobileNet model produced low angle errors when compared to the MobileNetV3 model. This also justifies that the \textbf{long-range dependencies learned by the self-attention modules are necessary for image tilt detection}. We also trained the ResNet-50 model as a baseline to validate the effectiveness of our training pipeline. Though, the ResNet-50 model outperforms both the MobileNetV3 and SA-MobileNet, the improvement is marginal despite the huge difference in the number of parameters between the ResNet-50 and the light-weight models. 

\textbf{Comparison with previous works}: We also present comparison of our training approach with previous works for this problem in Table \ref{tab:previousworks}. Over the past few years many different methods have been proposed to tackle this problem but they have high angle errors or low accuracies. ResNet-50 baseline model gives the lowest angle error on the test dataset to give state-of-the-art results. However, the ResNet-50 network cannot be deployed on low-resource devices because it has over 25 million parameters. But the proposed Self-Attention MobileNet has around 4 million parameters with low-latency values and state-of-the-art results for mobile devices that makes it ideal to be deployed in smartphones for real-time inferences. 
\section{Conclusion and Future Work}
We present a novel neural network model for mobile/IoT devices powered with the Self-Attention Modules. We also present highly accurate results on the task of Image Tilt Detection and are able to correct image orientations on smartphone in real-time. Our proposed training approach is also very simple but effective for this task. In the future work, we can use a dynamic value of the narrow interval, $I$, which can be unique for each image within limit. 

We will also validate the effectiveness of Self-Attention MobileNet on other downstream tasks like image classification, object detection and image segmentation applications for mobile devices.
\bibliography{egbib}
\end{document}